# A computer program for simulating time travel and a possible 'solution' for the grandfather paradox

Doron Friedman, The Interdisciplinary Center, Herzliya, Israel

doronf@idc.ac.il

**Abstract**

While the possibility of time travel in physics is still debated, the explosive growth of virtual-reality simulations opens up new possibilities to rigorously explore such time travel and its consequences in the digital domain. Here we provide a computational model of time travel and a computer program that allows exploring digital time travel. In order to explain our method we formalize a simplified version of the famous grandfather paradox, show how the system can allow the participant to go back in time, try to kill their ancestors before they were born, and experience the consequences. The system has even come up with scenarios that can be considered consistent "solutions" of the grandfather paradox. We discuss the conditions for digital time travel, which indicate that it has a large number of practical applications.

1. Introduction

In the principal paradox of time travel a person travels back in time and kills his grandfather before the grandfather meets the time traveler's grandmother. As a consequence, one of the traveler's parents, and therefore the traveler himself, would never have been born. This would imply that the traveller could not have travelled back in time, which means that the grandfather would still be alive, which now makes it possible for the traveler to be born, travel back in time and kill his grandfather, hence a paradox.

In this paper we describe a computer program that allows us to interactively explore the consequences of "changing the past" in a narrative. Our system uses well-known techniques from automated reasoning to compute the specific consequences of modifying history. The contribution of this paper is in the application of automated reasoning to the experience of time travel and in the implementation of this method in a computer program. We provide the details showing how our program was used to interactively explore the consequences of time travel while maintaining consistency in the context of a simplified version of the grandfather paradox. Furthermore, we show how the program was used to proactively resolve the paradox by suggesting transformations of the story line, resulting in some narratives that may be considered valid solutions to the famous paradox. We explain the difference between our time travel approach and other types of digital simulations. Finally, we explicate the conditions required for digital time travel, which indicate that the method can be applied in many useful domains.

Time travel is discussed in physics, philosophy, and popular culture. In this paper we deliberately avoid the discussion of time travel in physics, and try to capture the notions of time, causality, and time travel as they are perceived by laymen, assisted by insights provided by analytic philosophers. An early discussion of time travel in philosophy in contemporary times has been provided by Lewis (Lewis 1976). His suggestion for resolving the paradox rests on the hypothesis that if time travel is possible then some mundane event will always happen to prevent paradoxes from taking place. Horwich (Horwich 1975) criticized this view by pointing out that if travel to the local past is allowed then there would be countless attempts to initiate self-defeating causal chains, and it would be highly improbable that all of them would be avoided by mundane events such as slipping over a banana peel. Further investigations of time travel in philosophy lead to discussions of causality, identity, and other metaphysical issues (Dowe 2000, Grey 1999, Smith 1997). Our goal in this computer program is to borrow from these discussion and provide a virtual time travel experience that would be consistent with some popular notions of time travel. Nevertheless, we hope that future work based on our formalization of time travel in a computer program may contribute to the philosophical discussion.

Unlike the physical world, the digital sphere allows you to "go back in time", make a change, and observe the consequences. This can be as simple as the "undo" function in a word-processing program, or take place in the context of a rich experience such as in a virtual-reality simulation. For example, consider a video game in which you can be upset at the way things turned out at time *t*, restart the game and continue playing from a previous time $t'$, $t' < t$. Typically, in such cases whatever happened between $t'$ and $t$ is lost forever, and the simulation restarts. Our approach is different; it allows the participant to re-experience the history that took place between $t'$ and $t$. The history will be repeated, except for local changes caused by the participant's actions in the "second time around". The difference between our approach and typical simulations will be refined in Section 4, after providing the details regarding our method for virtual time travel.

Here we have focused on time traveling to the past. Time travel into the future is possible in physics (just hop on a spaceship that flies at almost the speed of light), but digital time travel to the future is still interesting to explore and may be useful in some scenarios. For example, meeting your older self in virtual reality (VR) was found to increase subjects' saving behavior (Hershfield et al 2011).

## 2. Method

Our implementation is based on two levels of abstraction: **logic** and **history**. The first level takes care of the automated reasoning mechanism, intended to maintain the logical consistency of the narrative. The second level of abstraction in necessary for the application level to present a time travel experience to the user. Automated processes convert application level assumptions and events into logical facts and constraints and vice versa.

**2.1 Logic**

Recording and replaying data is typically straightforward in the digital sphere. The main challenge in providing a digital time-travel experience is in automatically tracking causality chains. Fortunately, there are abundant formalizations and techniques in computational logic and artificial intelligence that can be utilized for tracking causality. We opted using Boolean constraint propagation; this is a straightforward mechanism that we have used in the past for automated reasoning in the context of VR applications.

The entities in this layer are terms, facts, and constraints. The terms in our grandfather paradox domain are mainly people and actions (supplementary text S.1). Facts are functions or actions applied to other terms. Exactly one of the terms needs to denote the current time to which the fact applies. At any given moment facts are assigned one of three truth values: **True**, **False**, or **Unknown**. The facts are treated as propositional; each fact is an atomic literal.

Constraints are logical operators applied to one or more facts; the constraints required for our domain are the Boolean operators: $\neg, \wedge, \vee$, and $\rightarrow$. We refer to these as constraints because the automated deduction process is based on constraint propagation.

The Boolean constraint propagation network is integrated with a truth maintenance system (TMS) (Doyle 1979, McAllester 1990) which records justifications for all derived values. The unit clauses in the reasoning database are treated as premises, which can be asserted or retracted by changing or removing the value of the corresponding node. The non-unit clauses (corresponding to constraint boxes in the network) cannot be changed. Finally, our reasoning engine also includes pattern-directed invocation (Rich & Feldman 1992): a procedure is automatically executed whenever a term matching a given pattern is created. In our case, pattern-directed invocation is used to implement quantifiers ($\forall$ and $\exists$).

**2.2 History**

A history is a sequence of state-action pairs denoted by: $h = ((s_0,a_0),(s_1,a_1),…,(s_n,a_n))$. The indexes $0,…,n$ correspond to the terms $0,…,n$ in the logical level of abstraction, which of course denote specific times.

A state is, in general, a collection of value assignments to a predefined set of variables $s = \{v_1,v_2,..,v_m\}$. In our domain the only variables are Boolean and reflect the existence of the actors in the scenario in the particular time. Thus, for each person $p$ in the application, there is one state variable $v_p$, which can be True, False, or Unknown, and indicates whether the person is alive or not at a given state.

For each action in the history level there is a corresponding fact in the logical level. The variable $v_p$ in state $s_t$ corresponds to the fact *exists(p,t)* in the logical level. Facts are generated dynamically as the narrative unfolds, i.e., whenever a new action is introduced to the history.

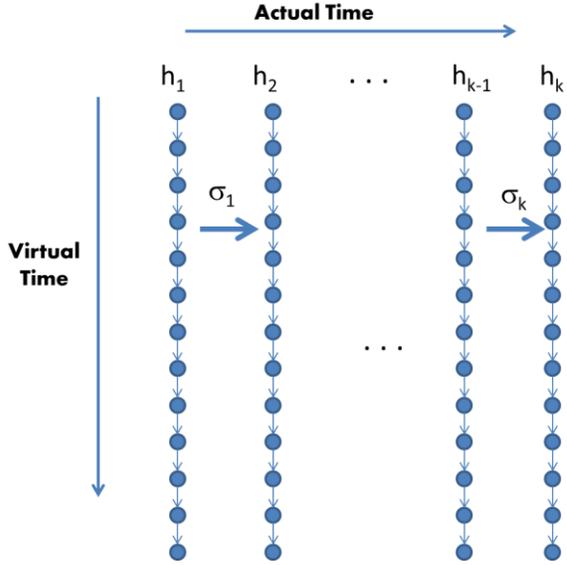

**Fig. 1:** A history is a sequence over virtual time. Histories can change by transformations ($\sigma_1,\ldots \sigma_k$) over actual time.

A history $h$ can be changed by adding or removing actions; this is written in the form: $h' = \sigma\,(h, A_h, A_{h'})$, where $A_h$ is the set of actions to remove from $h$, $A_{h'}$ is the set of actions to add to $h'$, and $\sigma$ is a transformation that changes $h$ into $h'$. The series of histories is generated through interaction with the participant over *actual time*, which is orthogonal to the *virtual time* of each history (Fig. 1). While we allow time travel along the virtual time dimension, actual time only moves in one orthogonal direction (at least until the physicists might come up with a physical time machine).

Given two histories $h_1$ and $h_2$, we can say that $h_1 \subseteq h_2$ if and only if the sequence of actions in $h_1$ is a subset of the sequence of actions in $h_2$ (when null actions are ignored). Typically, the series of histories will evolve by adding actions (since the simulation is interactive, and the participants take actions). Thus, as we move along the actual time histories become more specific, and each new history contains the previous history. If a transformation $\sigma$ s.t. $h_{k+1} = \sigma(h_k, A_1, A_2)$ does not satisfy $h_k \subseteq h_{k+1}$ then it is said to be a *strong transformation*. A strong transformation is one that transforms history by changing one of the things that already happened, rather than by just adding new actions. Strong transformations can be understood in the context of a multiple-world interpretation (Deutsch & Lockwood 1994). Before strong transformations are introduced, we can assume that there is one series of events, and information that is missing in the first histories is revealed over (actual) time. As soon as we have an action removed from a history, we can assume that this action took place in one universe and did not take place in another parallel universe.

Typically, actions are accumulated as the narrative unfolds, but actions can also be replaced or removed. Whenever a transformation is applied the resulting history needs to be consistent, i.e., the fact that we add an action $a_t$ poses additional logical constraints, and we require that both previous constraints and the new constraints will hold after the transformation. We allow a transformation to add a set of actions simultaneously, since it is possible that adding a single action would cause a contradiction, whereas adding two actions or more would allow for a consistent history.

### 3. Results: The main example

We formalize a simple version of the famous grandfather paradox (Horwich 1975, Lewis 1976). The example is kept deliberately as simple as possible for clarity. The example is constructed and explained in way that should facilitate communicating our ideas to a non-technical audience, and some of the technical information is left to the supplementary online material. We ignore issues of gender and sexual reproduction. We maintain the essence of the paradox: a son `S` goes back in time to kill his father `F` (rather than his grandparent) before `F` gives birth to `S`. This "toy" example does not require a history of over ten actions to illustrate the paradox and its "solutions". When formalized, the scenario includes a few hundred facts and constraints; the scenario of the "solution" to the paradox that we describe below (Fig. 6) includes 229 facts and 344 constraints. This "toy" example serves to explain our method and testify to its validity. Future work (in progress) will show how our method scales to a richer VR scenario.

### 3.1 Formalizing the paradox

### 3.1.1 Initialization and generic setup

Our example has only three possible actions: begetting, killing, and traveling in time. The precondition for the action `A beget B` is that `A` is alive and the post-condition is that `B` is alive. The preconditions for `A kill B` is that both `A` and `B` are alive, and the post-condition is that `B` is not alive. Time travel is modeled as two interlinked actions, depart and arrive. A person has to exist in order to depart (precondition) and after he arrives (post-condition).

A notorious challenge when modeling actions over time is the frame problem (McCarthy & Hayes 1968). When modeling a dynamic domain it is natural to specify the changes that take place whenever an action is taken. However, the system also needs to be able to deduce that all the rest has not changed. Since our domain is very small we have taken the naïve approach: we explicitly specify for each action whether it changes the state variables in the world. In this case the state variables are only required to track whether a specific person exists or not, so this solution is reasonable. Extension of the work presented here needs will reveal whether this simple method is scalable or whether any of the many formalisms suggested for overcoming the frame problem (e.g., (Hanks & McDermott 1986, Reiter 1991, Scherl & Levesque 2003, Schubert 1990, Shanahan 1997, Shoham 1987)) may be more appropriate.

We introduce two constraints: **remains** and **appears**. For each person we define that he always remains at time *t+1* if he existed in time *t*, and that he never appears at time *t+1* if he did not exist at time *t*. We call these the continuity-of-existence and the continuity-of-non-existence rules. They are implemented automatically using pattern-directed invocation. We introduce these constraints whenever a new person is entered into the simulation. Thus, at this stage there are no actions that involve this person. Whenever a new action involving that person will be introduced the system will check whether the default assumptions about remaining and appearing need to be revised. Thus, it is a type of default reasoning that is practical and simple (see Section S.2 of the supplementary online material for details).

There are two ways to explain our approach. So far we have adopted the formal way, which is more convenient from an algorithmic point of view would adopt a meta-temporal viewpoint ("the view from nowhen" (Price 1996)). In this view we are outside the timeline and at any moment can observe a complete history from beginning to end, or even multiple histories, in the case of a multiple world interpretation. However, note that in order to provide a time travel experience the participant needs to be embedded inside the timeline, and thus be provided with only a limited view of the present. The implementation in our program is straightforward: the participant is at any moment (of real time) in a specific state (of virtual time). Only the current values of the state variables are passed into the rendering engine that describes the environment. In our program described in this paper the user sees a textual description of the world. In general, the state of the world can be rendered using a more immersive virtual reality.

As the real time moves along, so does the virtual time progresses. The participant steps from one state to the consequent state, in the order of the virtual time. The exception is time travel; in this case of course the order of states visited by the participant is violated, and she is transferred to an earlier or later state according to the setup of the `time machine'. The implementation is the same; the `current time' can skip to a previous state or to a state in the future, and that state information is sent to the rendering engine.

Our example includes two people: a time traveler `S` and his parent `F`. Already, we have a chicken and egg problem; we would like to formalize the fact that people are born to their parents, but this raises a question: where did the first person come from? Since modeling evolution is out of the scope of this paper, we introduce a person called `F` to exist in the world from the beginning, by adding a corresponding fact at the logic layer. Based on the continuity-of-existence rule mentioned above, the system deduces that `F` exists throughout the whole history (Section S.2).

In our example the time traveler, `S`, is a user of the computer program, and the narrative will unfold from `S`'s the point of view. In order to begin the scenario the system introduces the action whereby `F` gives birth to `S`. The system has some knowledge about the nature of giving birth – it requires that a person would exist to give birth and it also requires that a person would not exist before it is born (Section S.3).

Our program gives the user free choice to take different actions, but here we are interested with the exploration of the paradox. We will assume that the user plays along in order to explore the paradox. Thus, after being born, `S` goes back in time to some other point in time before he was born. At this point our system reports a logical contradiction.

When `F` begets `S` this introduces the precondition that `S` does not exist before the birth, whereas now he does exist, since he arrived via time travel. The contradiction thus involves a person already existing at the time they are born; this is impossible in our world but is of course made possible by time travel (Section S.4).

At this point we introduce three "metaphysical" models: $T_0$ – it is not allowed for a person to exist at the time they are born, $T_1$ – the old copy merges into the newly born copy so there is always a single copy of each person at most, and $T_2$ – every time a person goes back in time a new copy (clone) is generated, and multiple clones may co-exist in time; this version is popular in fiction. Even $T_0$ does not take out all the fun of time travel, since time travel is possible both into the future and into the remote past (Fig. 2). $T_1$ can be useful for some applications but it raises problems, e.g.: if a grown up person would travel back in time to the time they are born, $T_1$ would imply that a grown person has merged into a newly-born baby, which probably does not reflect our intention.

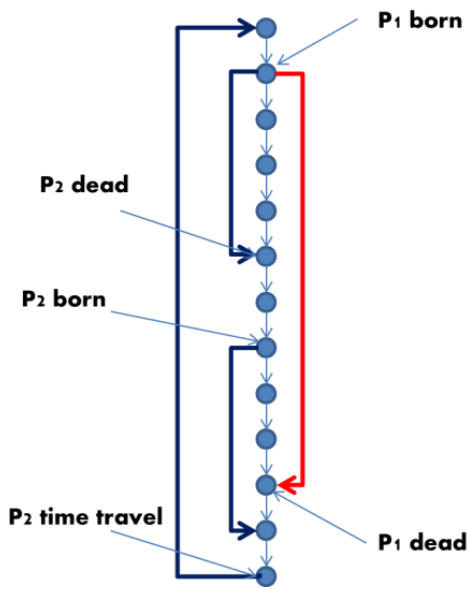

**Fig. 2**: A time line with two people: $P_1$ has a standard lifeline, and $P_2$ has a lifeline with one time travel to the past. This is consistent with model $T_0$, since P2's clone does not exist at his birth.

For the rest of the scenario we adopt model $T_2$, which means that following time travel `S` turns into a clone `S₁`, and they both co-exist (Section S.4). Thus, we resolve the contradiction by technically relaxing the constraint on identity; a time traveler and his clone are two separate entities. Thus, when `S` is born his clone `S₁` exists and witnesses the birth, but S is technically different than $S_1$, so the contradiction is resolved. Note that this is just a technical solution; S and $S_1$ are one in terms of a continuous phenomenal experience, and it is still the case that the son killed his parent. A person and their clone have the same personal identity in the same way that a child and the same person who is a grownup have the same personal identity (see (Lewis 1976) for a discussion on identity and time travel) (Section S.4).

We are now ready for the paradoxical murder to take place. The user (who is now the cloned son $S_1$) kills his father F. On the history timeline this happens before the father begets the user S (Section S.5). This is the parent paradox, our version of the principle paradox of time travel (Fig. 3).

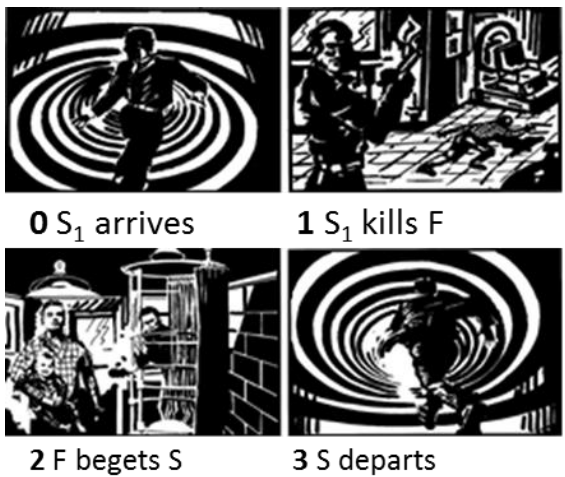

**Fig. 3**: An illustration of the parent paradox. The son S goes back in time, becomes the clone $S_1$, and kills his father F.

### 3.2. Exploring the paradox

The user (clone $S_1$) kills his parent F; this introduces the parent paradox and, indeed, our reasoning engine reports a contradiction: F is supposed to be dead because he was killed, but he is also supposed to be alive to beget S (Section S.8). First, our system allows the user to explore the implications of the paradox and to try to resolve it, by retracting the offending premises. The system reports the facts that are involved in the contradiction, and the user can decide that the actions that assigned these facts their truth values would not have happened. For example, the user can retract the murder, or alternatively retract the beget action, and `F` is no longer `S`'s father (Section S.9).

Rather than allowing the user to resolve paradoxes "manually", the system can proactively suggest transformations that resolve contradictions by adding new actions to the history automatically; some philosophers even suggest that our world may behave in such a way (recall the banana peels?;*4,7*). There are various approaches as to how the system can proactively try to resolve paradoxes. From a computational point of view this can be posed as a logical deduction problem, and the system can come up with all the solutions, if any, that are compatible with the logical constraints. However, the search for a solution is in principle computationally expensive (exponential). Moreover, if there is one possible solution for resolving the paradox then there would typically be many solutions. Thus, our goal is to quickly find the "best" solutions to the paradox.

Defining what the best solution is an interesting question. We may pose some constraints on the solutions, which would disqualify some of the logically acceptable solutions. For example, we would probably disqualify solutions that assume that the user has to do something very specific, thus overriding the user's free choice. In addition, we may want to sort the solutions in terms of how plausible they seem. For example, we can require the system to introduce a change such that the new history is as close as possible to the old one; this is the kind of "even with time travel you cannot escape your fate" narrative often presented in fiction, such as the feature film The Butterfly Effect[1]. Section S.10 provides a technical formalization of this intuition.

These methods for resolving paradoxes are generic. In the case of the parent paradox the system adopts a domain-specific strategy. Since our domain is simple we have augmented our system with some meta-knowledge. In our model world, there are two typical types of contradictions: either a person is supposed to exist at some point and he does not, or a person is supposed not to exist at some point and he does. In the first case, the system tries to introduce actions that cause the person to exist at the required point in time: either by having the person born or by making him arrive by time travel. In the second case the system tries to introduce actions that would "get rid" of the person at the required time, either by having him killed or having him leave by time travel. Such automatic attempts to resolve the contradiction can result in various solutions, and some of these solutions are discussed in more detail in the supplementary online material (Section S.11).

---

[1] http://www.imdb.com/title/tt0289879/

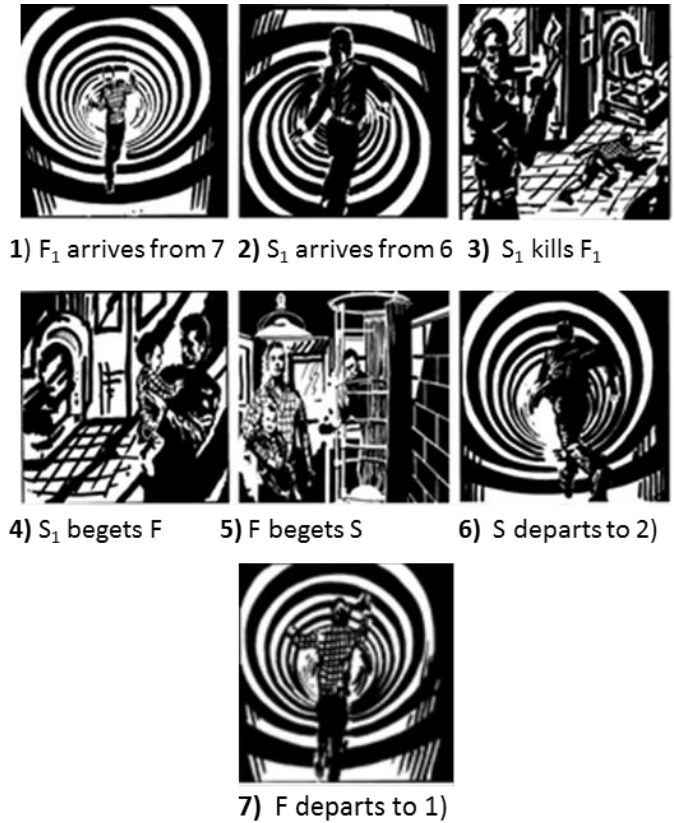

**Fig. 4**: F gives birth to S and goes back in time. S, right after being born, goes back in time as well, and kills his parent F. Right after killing his parent, though, S (actually the clone, $S_1$) gives birth to him. In this solution, therefore, S is F's son, F's father, and F's killer.

Here we provide two solutions of interest. Applying the deduction mechanism without constraints yields the first solution (Section S.11): the son goes back in time, kills his father, and then gives birth to his father, which would then give birth to him (Fig. 4). This solution results in an entertaining narrative, and in some sense it serves as a `solution' to the paradox, but it has a drawback. The system violates the time traveler's `free will' in the second time around, and forces the user ($S_1$) to beget F.

Therefore, we prefer another solution, which can be automatically deduced by the system and does not assume anything of the user. Figure 5 describes what we regard to be a simple yet novel solution to the parent paradox (which is also, of course, a solution to the grandfather paradox): the son goes back in time and kills his father. From the user's (son's) perspective this seems like a paradox. Only if the user waits long enough in the simulation he may discover how the paradox was resolved. Without the son's knowledge it turns out that the father actually traveled to the future to give birth to him and came back just before being killed. How were we able to avoid the paradox? `S` killed `F` **before** `F` gave birth to `S` in virtual time, but the murder happened **after** the birth with respect to S's personal time.

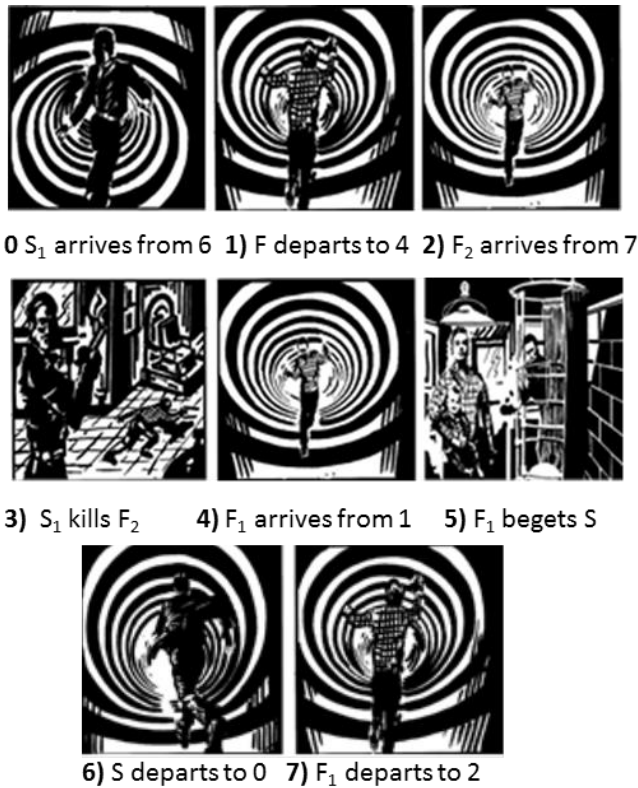

**Fig. 5**: An illustration of a solution of the parent paradox. F travels to the future, begets S, and travels back just in time to be killed by a clone of S who also went back in time.

## 4. Discussion

In this paper we attempted to capture the common-sense interpretation of time and causality, which has been claimed to be not only wrong but also logically inconsistent. Since we allow changing history, our interpretation is similar to Deutsch's interpretation of experiencing time travel in parallel universes through virtual reality (Deutsch 1997), but there are subtle differences; for example, in the first time around our time traveler does not need to meet his future clone. Most important, using our method, every assumption that is explicitly introduced into the system can be altered and the consequences can be further explored. Thus, our approach allows exploration of the consequences of various metaphysical assumptions, including ones that we believe to be true about the physical world.

Given that we can (1) record events, (2) define the causality relationships among the events, and (3) replay events, we can implement various forms of time travel. These three conditions are not met in the physical world (mostly, of course, the third one), but they can often be met in the digital sphere (here the main challenge is on the second condition, modeling and capturing causality). Thus, there are a large number of digital applications for which we can introduce time travel, e.g., for entertainment, training, and psychological interventions.

Now we can refine the differences between our approach and a "traditional" simulation. Obviously, in simulations in general you do not meet your former clones and do not interact with other time travelers. However, there is also an important logical difference. If a simulation is completely deterministic then the narrative should unfold in the same way as it does using our approach. However, many domains require non determinism for useful simulations. Even if the world is completely deterministic, we do not have a complete deterministic model of how people make decisions, individually or collectively. This is especially true when modeling people and historical events, but is also true in many other domains. Thus, our simulations would necessarily include idiosyncratic events, which would not repeat exactly as is in a typical simulation, and this is where our approach is required.

The main limitation of our approach is in formalizing causality in large-scale domains of knowledge; this is a notoriously difficult problem in artificial intelligence, and one of the main reasons that "classic" approaches have come out of favor. We expect our method, as described here, to scale to complete applications in some small well-defined domains, but large-scale implementations may require substantial changes.

Future work should correspond with the vast literature in artificial intelligence, logic, and philosophy regarding time, action, and causality. In this paper we have used relatively basic methods but future work may be bi-directional; first, theoretic insights from those fields may assist in refining our approach, scaling it, and reinforcing its foundations. Second, we hope that our practical approach will contribute to the theoretic discussions in those fields.

Our future work focuses on the exploration of our approach within an immersive virtual reality system. Such a system would require refinements and extensions to our method, and will serve to reaffirm its validity. Moreover, with the aid of an immersive system we hope to explore the psychological experience of virtual time travel; such exploration may be useful not only as a form of art and entertainment, but may serve as a useful tool for education, training, and psychological science and rehabilitation.

**Acknowledgments**

I wish to thank Rafael Malach, Mel Slater, and Thomas Metzinger for their feedback on this manuscript.

**Online Supplementary Material**

**S.1: Terms in the paradox domain**

For the paradox example our domain includes propositions made of the following terms:

1. P: people, specifically **p** (parent) and **s** (subject).

2. A: actions, including: **beget, kill, depart**, and **arrive**. Additionally, we include a no operation action **(nop)**.

3. T: time instances – positive integers denoting a specific time.

4. F: functions, including: **exists**, **remains**, and **appears.**

**S.2: Initialization and the frame problem**

A specification of a history includes the truth values of the state variables at the beginning of each state. The history is initialized to be empty, and the main actors, P and S, do not exist – this is marked as ? (unknown) at all times.

Table 1: The history (states and actions) in our domain upon initialization

| Time | P | S | Action |
|---|---|---|---|
| 0 | ? | ? | nop |
| 1 | ? | ? | nop |
| 2 | ? | ? | nop |
| 3 | ? | ? | nop |

Whenever an action is introduced in the history level a corresponding fact is automatically introduced into the logic level. At the initialization stage the application thus introduces the following facts into the logic level:

nop(0)

nop(1)

nop(2)

nop(3)

In order to introduce the father (p) we add an axiom to the logic level:

exist(p,0)

In our example, adding exists(p,0) results in the following facts, all initialized to Unknown:

exists(p,1)

exists(p,2)

exists(p,3)

remains(p,0)

remains(p,1)

remains(p,2)

remains(p,3)

appears(p,0)

appears(p,1)

appears(p,2)

appears(p,3)

The same rule now installs a set of constraints that require that all actions in the history do not affect the state of that person; this is always true since this only happens at the first time the person is introduced. In this case the rule installs the following constraints:

nop(0) $\rightarrow$ remains(p,0)

nop(1) $\rightarrow$ remains(p,1)

nop(2) $\rightarrow$ remains(p,2)

nop(3) $\rightarrow$ remains(p,3)

nop(0) $\rightarrow$ appears(p,0)

nop(1) $\rightarrow$ appears(p,1)

nop(2) $\rightarrow$ appears(p,2)

nop(3) $\rightarrow$ appears(p,3)

Every time a new constraint is introduced it immediately applied. In our case this sets the following facts to be True:

remains(p,0)

remains(p,1)

remains(p,2)

remains(p,3)

appears(p,0)

appears(p,1)

appears(p,2)

appears(p,3)

In addition, the rule installs the following constraints; these maintain the continuity and discontinuity of people's existence, in both temporal directions.

exists(p,0) ^ remains(p,0) → exists(p,1)

exists(p,1) ^ remains(p,1) → exists(p,2)

exists(p,2) ^ remains(p,2) → exists(p,3)

exists(p,3) ^ ¬appears(p,2) → exists(p,2)

exists(p,2) ^ ¬appears(p,1) → exists(p,1)

exists(p,1) ^ ¬appears(p,1) → exists(p,0)

¬exists(p,0) ^ ¬appears(p,0) → ¬exists(p,1)

¬exists(p,1) ^ ¬appears(p,1) → ¬exists(p,2)

¬exists(p,2) ^ ¬appears(p,2) → ¬exists(p,3)

¬exists(p,3) ^ remains(p,2) → ¬exists(p,2)

¬exists(p,2) ^ remains(p,1) → ¬exists(p,1)

¬exists(p,1) ^ remains(p,0) → ¬exists(p,0)

These constraints are evaluated and result in adding the following facts to the system:

exists(p,1)

exists(p,2)

exists(p,3)

These facts and constraints together provide the solution to the frame problem, in our limited domain. We can prove, for example, that a person exists until he is killed or he departs via time travel; this is guaranteed by the remains facts.

Following these steps, the history is updated as follows:

Table 2: The history after adding an axiom specifying that P exists at time 0, and after constraint propagation.

| Time | P | S | Action |
|------|---|---|--------|
| 0 | + | ? | nop |
| 1 | + | ? | nop |
| 2 | + | ? | nop |
| 3 | + | ? | nop |

## S.3

The first step in the narrative is the birth of the son s. We will introduce the begetting of the son s by the parent p as the third action, which is the action that is applied after the state $s_2$. This is where non-monotonic reasoning is applied; the previous action is discarded by resetting the fact nop(2) to False and a new fact is introduced to be True: beget(p,s,2).

For each type of action there are specific constraints that need to be installed; this is a generic mechanism that can be implemented with pattern-directed invocation. These constraints reflect the semantic of the action. In our case the constraints are:

beget(p,s,2) → remains(p,2)

beget(p,s,2) → ¬appears(p,2)

beget(p,s,2) → appears(s,2)

beget(p,s,2) → exists(p,2)

beget(p,s,2) → ¬exists(s,2)

appears(s,2) → exists(p,3)

The history is updated as follows:

| Time | P | S | Action |
|------|---|---|--------|
| 0 | + | - | Nop |
| 1 | + | - | Nop |
| 2 | + | - | beget(p,s) |

| 3 | + | + | Nop |

In our interactive application this is actually the beginning. The user is s. The user would typically not see the "god's view" of the whole history, but only the current time. So the time is 3, and the user is provided a description of the state of the world. In our application the description only includes a textual description of the current time, and of the fact that he and his parent, p, exist. In general we could envision a virtual reality environment that renders this state as a complete audio-visual experience.

Next, we introduce the time travel action from time 3 to time 0. This is done by adding two actions, resulting in two facts: depart(s,0) and arrive(s,3) (the second parameter in both actions denotes the destination and source of the time travel, correspondingly). The facts are set to be True. The introduction of these actions also results in the following constraints:

depart(s,0) → ¬remains(s,3)

depart(s,0) → ¬appears(s,3)

depart(s,0) → exists(s,3)

arrive(s,3) → remains(s,0)

arrive(s,3) → ¬appears(s,0)

arrive(s,3) → ¬exists(s,0)

The history is:

| Time | P | S | Action |
|---|---|---|---|
| 0 | + | - | arrive(s,3) |
| 1 | + | + | Nop |
| 2 | + | + | beget(p,s) |
| 3 | + | + | depart(s,0) |

### S.4: The time-travel contradiction

After adding the time travel, the system reports a logical contradiction: s has to exist at time 1 since he arrived there by time travel, but he also should not exist at that time because he was born later, between times 2 and 3!

As explained in the main text, there are three time travel models: $T_0$, $T_1$, and $T_2$. In addition, the program can be run in three modes:

$R_1$: The system prevents the user from doing anything that results in a contradiction

$R_2$: The system allows the user to interactively try to resolve contradictions

$R_3$: The system automatically tries to resolve contradictions.

In all modes the system uses the deduction chain that led to the contradiction.

exists(s,1)

arrive(s,0) $\rightarrow$ exists(s,1)

arrive(s,0) is True: premise

$\neg$ exists(s,1)

$\neg$ exists(s,2) $\wedge$ remains(s,1) $\rightarrow$ exists(s,1) TBD: make sure we have it above

beget(p,s,2) $\rightarrow$ $\neg$ exists(s,2)

beget(p,s,2) is True: premise

nop(1) $\rightarrow$ remains(s,1) TBD: make sure this is consistent

nop(1) is True: premise

Specifically, the logic layer reports the premises. Usually these would be actions; in this case there are indeed three facts that are premises and correspond to the three actions in times 0,1, and 2.

In model T0 this is indeed a contradiction. If the program is running in mode R1 the user will not be allowed to go back in time to the past. This is very limiting. Thus, if in option $R_1$ the contradiction is delayed until the last moment. In our case the contradiction involves 3 actions, at times 0,1, and 2. The program can keep this contradiction in a buffer and wait until time 2, to see if the contradiction is resolved by then. For example, the user may choose to travel in time again, and would thus not exist at the time he is born.

TBD: prove that dealing with delayed contradictions works.

In mode $R_1$ the user may be given a choice to manually overcome the contradiction. This is typically used for debugging the application, rather than in the application itself. The user is given the set of premises that serve as the roots in the deduction tree that leads to the contradiction, and the user is required to retract one of the premises.

The premises in this case are:

arrive(s,0) #TBD check syntax

nop(1)

beget(p,s,2)

If the arrive is retracted a mechanism makes sure the departure is also retracted. This can be done by logic but we prefer to do it hardcoded. TBD. This is equivalent to avoiding the time travel.

In general when an action is cancelled it is replaced by a nop. Since the user asks to replace a nop, he is asked for a new action. Here the user can add an action that would cause the user not to exist after it, i.e. killing the user or time traveling.

Finally, the last option is remove the begetting of the user; typically the application would not allow this to happen.

In mode $R_2$ the system automatically tried to resolve the contradiction. There are several ways of doing this, but in the domain of the parent paradox the most useful is a heuristic rule. This would be useful later for resolving the main paradox, so we provide here the details.

Note that the typical contradictions in the parent domain take place when a person is supposed to exist for one reason and is supposed not to exist for another reason. The heuristic for resolving such constraints is as follows.

$A_e \leftarrow$ set of actions that require p to exist

$A_n \leftarrow$ set of actions that require p not to exist

$T(A) \leftarrow$ set of times of the actions in A

$t_{en} \leftarrow min(T(A_e))$

$t_{ex} \leftarrow max(T(A_e))$

$t_{an} \leftarrow min(T(A_n))$

$t_{ax} \leftarrow max(T(A_n))$

if $t_{ex} < t_{an}$ then

    find t, $t_{ex} < t < t_{an}$ s.t. nop(t) is True

    replace the nop action by an action that eliminates p

else if $t_{en} > t_{ax}$ then

    find t, $t_{ax} < t < t_{en}$ s.t. nop(t) is True

    replace the nop action by an action that introduces p

endif

Note that the algorithm does not always apply but practically it will most of the times, and this is also the case in our examples. If the domain-specific heuristics fail we resort to general techniques.

Note that if there are no nops in the middle we can introduce a new state followed by a nop. We have assumed that the resolution of time is infinite. The details are TBD.

In our case the algorithm needs to replace nop(1) with an action that eliminates s. This could be a time travel or killing s. The first kind of makes sense – your time machine ejects you to another unexpected time to avoid contradictions. For example:

| Time | P | S | Action |
|------|---|---|--------|
| 0 | + | - | arrive(s,3) |
| 1 | + | + | depart(s,4) |
| 2 | + | - | beget(p,s) |
| 3 | + | + | depart(s,0) |
| 4 | + | - | arrive(s,1) |

The second type of solution is more dramatic:

| Time | P | S | Action |
|------|---|---|--------|
| 0 | + | - | arrive(s,3) |
| 1 | + | + | Kill(p,s) |
| 2 | + | - | beget(p,s) |
| 3 | + | + | depart(s,0) |

In this narrative s goes back and is killed by his parent, just before the parent gives birth to s. Game over.

In general there are infinitely many ways for the system to automatically resolve the contradiction. The general criteria is to find all solutions that do not require extending the time line. There are at most a finite number of such solutions. If there is no such solution the system gradually extends the time line until a solution is found. We also weight the changes. For example, adding a time travel is prepared over killing the user; not for ethical reasons, but because we don't want to terminate the application.

TBD: this algorithm above is infinite. Can we prove a reasonable bound? Then we can sort all legal solutions. In the example above if we prefer shorter solutions then the user would be killed.

In model $T_1$ we allow only one copy of each person to exist. We remove the preconditions that a person does not exist before he is born or arrives. [TBD – do I have it for arrives?]. TBD do we need the appears facts at all?

The challenge is that there is a very twisted personal time – is it consistent at all? TBD

We adopt model $T_2$, which means that following time travel s turns into a clone $s_1$, and they both co-exist.

| Time | P | S | $S_1$ | Action |
|---|---|---|---|---|
| 0 | + | - | - | arrive(s1,3) |
| 1 | + | - | + | Nop |
| 2 | + | - | + | beget(p,s) |
| 3 | + | + | + | depart(s,0) |

A clone is also generated automatically in the case of time travel to the future.

It is easy to see that this avoids the contradiction: exists(s,1) is now False. The challenge is to show that this technical solution involving clones retains our intuitions about personal identities. What are these intuitions? Before time travel, it would be:

1. p is born once, say at t1

2. p dies once, say at t2

3. p is alive for each t, t1 ≤ t ≤ t2

4. p is not alive for each t, t < t1

5. p is not alive for each t, t > t2

Note that this is the continuity of existence assumption for each object, not only for people.

If we introduce time travel, axioms 3-5 no longer hold. We replace this with the following set of requirements:

1. There is a finite set of intervals $I_1=[t_{11},t_{12}]$, …, $I_n=[t_{n1},t_{n2}]$ in which p is alive.

2. there is exactly one k s.t. p is born in tk1.

3. there is exactly one j s.t. p dies in tj2.

4. For every l, l~=k, 1 <= l <= n, for every tl1 there is an action arrive(p,tl1,t') at this time. The arrival is associated with a departure time t', s.t. there is an index i s.t. ti2 = t', and i~=j.

5. For every l, l~=j, 1 <= l <= n, for every tl2 there is an action depart(p,tl2,t') at this time. The departure is associated with an arrival time t', s.t. there is an index i s.t. ti1 = t', and i~=k.

TBD: prove that these conditions cover overlap.

TBD: so what do we do with these definitions? Need to prove that they hold. Using the continuity of existence and in-existence rules. But still need to make sure no one lives twice. So add axioms. And move the standard people, before time travel, to an earlier section.

**S.5: The parent paradox**

Killing actions result in the following constraints and propositions:

kill(a,b,t) $\rightarrow$ remains (a,t)

kill(a,b,t) $\rightarrow$ ¬appears (a,t)

kill(a,b,t) $\rightarrow$ ¬remains (b,t)

kill(a,b,t) $\rightarrow$ ¬appears (b,t)

kill(a,b,t) $\rightarrow$ exists (a,t)

kill(a,b,t) $\rightarrow$ ¬exists (b,t+1)

We add kill (s1,p,1) and set it to True. The result is a contradiction.

| Time | P | S | $S_1$ | Action |
|---|---|---|---|---|
| 0 | + | - | - | arrive(s1,3) |
| 1 | + | - | + | kill(s1,p) |
| 2 | X | - | + | beget(p,s) |
| 3 | + | + | + | depart(s,0) |

Note: X is where the contradiction is. The parent has to exist to beget s, but he was just killed by s1.

**S7**

The condition for a preservation of identity with the possibility of time travel is that there would be a continuous line. @@TBD formalize. Moreover, there should be a continuity of the properties of the object, or the attributes of the person. @@TBD formalize. In our case there are no attributes, the only attribute is whether a person is dead or alive. If we had attributes, clearly the identity can be preserved under time travel and in this way the clones all have a single identity.

**S.7. Resolving the paradox: The naïve version**

In knowledge based systems we can often let the user decide what option to retract (see above) – this is useful for debug.

There are two facts involved in the contradiction. One of them was introduced by the user, and the other in the initialization. In principle the system can retract the initialization fact – the begetting of the user, but then the user would not exist. In principle the system can be smart enough to decide that the user exists for another reason, and avoid the paradox by having the user not be the son of the one he kills. Not sure if this is interesting and worth the effort.

**S.8. Proactive reasoning**

The first principle is that we do not retract premises. Premises may be of three types: a) world or application knowledge (in our domain we do not have any – only constraints), b) application specific initialization of narrative, and c) user actions. In general it may be possible to change some of these without changing the "spirit of the story", but this is beyond our scope.

By induction the history is consistent at any given time t, so the contradiction is always attributed, at least partially, to the most recently introduced action. In general, we discuss this in terms of transformations rather than actions; recall that transformations may involve removing some actions as well as adding others.

A generic proactive reasoning method, intended to keep the changed history as close as possible to the original one, can be written formally as follows. Given a transformation $h_2 = \sigma(h_1, A_1, A_2)$ s.t. $h_2$ is illegal, find sets of actions $A_3, A_4$ s.t. $h_3 = \sigma(h_2, A_3, A_4)$ is consistent and for each possible $h \in H$ $|h_3-h_1| \leq |h-h_1|$. The distance $|h_1-h_2|$ is typically application dependent. One criteria for this metric is that some actions are less common than others. For example, adding a beget action in our domain may be more likely than adding a killing action. There may be additional constraints on $A_3$ and $A_4$; e.g., the system would typically not introduce actions taken by the user (so as not to violate his free will or his memory of his own actions).

We have introduced our domain-specific algorithm to overcome contradictions in Section X. TBD: could be generalized. It is applied here in a very similar way, with interesting results.

Here is the deduction chain leading to the contradiction:

exists(p,2)

beget(p,2) → exists(p,2)

beget(p,2) is True: premise

but also

¬exists(p,2)

kill(s1,p,1) → ¬exists(p,2)

kill(s1,p,1) is True: premise

The algorithm is applied. The premise for the non existence of p is at time 1 and the premise for its existence is from time 2. So, in this case, the algorithm needs to add an action between 1 and 2, which makes p appear. Here is a possible solution:

| Time | P | S | $S_1$ | Action |
|---|---|---|---|---|
| 0 | + | - | - | arrive(s1,3) |
| 1 | + | - | + | kill(s1,p) |
| 1.5 | - | - | + | beget(s1,p) |
| 2 | + | - | + | beget(p,s) |
| 3 | + | + | + | depart(s,0) |

This could have been a solution unless we have made sure each person exists only once. So the kill action at time 1 installs not exists for all time > 1, and similarly the new beget at time 1.5 installs a not exists at all times < 1.5. So this solution, which results in p having tow lifelines, is invalid.

TBD: adding an action in the middle 1.5 – what happens to continuity?

$S_1$ killed F and the system decided that he then gave birth to him again! Here we display the whole history, for the reader's convenience. This is an interesting twist, but it has two problems. The first is that we do not want the system to take proactive actions on behalf of the user. If we had another user they could beget F, and that would make more sense.

The other problem is an identity problem, and a very interesting one! Incidentally, this solution to the paradox and its importance was discussed by philosophers (see (Smith 1997), counterfactual B, p. 372). We see that S went back in time and killed F, and F is S's father. How did the system avoid the parent paradox? The problem is, of course, that we now have two separate life threads, both called F, and the F that was killed by S is not F's father.

**S.9 A solution to the paradox**

| Time | P | P$_1$ | S | S$_1$ | Action |
|---|---|---|---|---|---|
| -1 | - | - | - | - | arrive(p1,4) |
| 0 | - | + | - | - | arrive(s1,3) |
| 1 | - | + | - | + | kill(s1,p1) |
| 1.5 | - | - | - | + | beget(s1,p) |
| 2 | + | - | - | + | beget(p,s) |
| 3 | + | - | + | + | depart(s,0) |
| 4 | + | - | - | - | depart(p,-1) |

How do we reach this? the system needs another heuristic to generate time travel .

TBD: can add an arrival instead of the beget! Is that a new solution? Or is that how the system arrives at the solution below?

| Time | P | P$_1$ | P$_2$ | S | S$_1$ | Action |
|---|---|---|---|---|---|---|
| 0 | + | - | - | - | - | arrive(s1,3) |
| 0.5 | + | - | - | - | + | depart(p,1.5) |
| 0.75 | - | - | + | - | + | arrive(p2,4) |
| 1 | - | - | + | - | + | kill(s1,p2) |
| 1.5 | - | - | - | - | + | arrive(p1,0.5) |
| 2 | - | + | - | - | + | beget(p1,s) |
| 3 | - | + | - | + | + | depart(s,0) |
| 4 | - | + | - | - | + | depart(p1,0.75) |

The solution needs to add time travel as well.